\documentclass[english, 10pt, twocolumn, twoside]{IEEEtran}
\usepackage{threeparttable}
\usepackage{mathrsfs}

\usepackage{cite}

\ifCLASSINFOpdf
\else
\fi
\usepackage{graphics}

\usepackage{subfigure}
\usepackage[cmex10]{amsmath}
\usepackage{amsfonts}
\usepackage{cases}
\usepackage{epsfig}
\usepackage{url}

\usepackage{algorithm}
\usepackage{algorithmic}

\usepackage{array}
\usepackage{amsmath}

\usepackage{multirow}
\usepackage{bigstrut}

\usepackage{bm}
\usepackage{amsmath}
\usepackage{amssymb}
\usepackage{mathrsfs}
\usepackage{color}

\usepackage{booktabs}
\usepackage{multirow}
\usepackage{amsmath}
\usepackage[bottom]{footmisc}
\usepackage{bbm}
\usepackage{hyperref}
\hypersetup{
	colorlinks=true
}

\def\etal{\emph{et al}.~}

\newcommand {\bt}[1]{\bf{#1.}\normalfont}

\begin{document}

%
\title{Unsupervised HDR Image and Video Tone Mapping via Contrastive Learning}
%
%
%

\author{Cong Cao, Huanjing Yue, \textit{Member, IEEE}, Xin Liu, \textit{Senior Member, IEEE,} Jingyu Yang, \textit{Senior Member, IEEE}
\thanks{This work was supported in part by the National Natural Science
Foundation of China under Grants 62072331, 62231018, and 62171309.
C. Cao (caocong\_123@tju.edu.cn), H. Yue (huanjing.yue@tju.edu.cn), and J. Yang (yjy@tju.edu.cn) are with the School of Electrical and Information Engineering, Tianjin University, China. X. Liu (linuxsino@gmail.com) is with the School of Electrical and Information Engineering, Tianjin University, China, and Computer Vision and Pattern Recognition Laboratory, School of Engineering Science, Lappeenranta-Lahti University of Technology LUT, Lappeenranta, Finland. \textit{Corresponding author: Huanjing Yue.}}}

%
%

\markboth{}%
{}
%



\maketitle

\begin{abstract}
Capturing high dynamic range (HDR) images (videos) is attractive because it can reveal the details in both dark and bright regions. Since the mainstream screens only support low dynamic range (LDR) content, tone mapping algorithm is required to compress the dynamic range of HDR images (videos). Although image tone mapping has been widely explored, video tone mapping is lagging behind, especially for the deep-learning-based methods, due to the lack of HDR-LDR video pairs. In this work, we propose a unified framework (IVTMNet) for unsupervised image and video tone mapping. To improve unsupervised training, we propose domain and instance based contrastive learning loss. Instead of using a universal feature extractor, such as VGG to extract the features for similarity measurement, we propose a novel latent code, which is an aggregation of the brightness and contrast of extracted features, to measure the similarity of different pairs. We totally construct two negative pairs and three positive pairs to constrain the latent codes of tone mapped results. For the network structure, we propose a spatial-feature-enhanced (SFE) module to enable information exchange and transformation of nonlocal regions. For video tone mapping, we propose a temporal-feature-replaced (TFR) module to efficiently utilize the temporal correlation and improve the temporal consistency of video tone-mapped results. We construct a large-scale unpaired HDR-LDR video dataset to facilitate the unsupervised training process for video tone mapping. Experimental results demonstrate that our method outperforms state-of-the-art image and video tone mapping methods. Our code and dataset are available at \href{https://github.com/cao-cong/UnCLTMO}{https://github.com/cao-cong/UnCLTMO}.
\end{abstract}

\begin{IEEEkeywords}
Image and video tone mapping, contrastive learning, video tone mapping dataset
\end{IEEEkeywords}

%
\IEEEpeerreviewmaketitle

\section{Introduction}
%
%
%
%
In recent years, high dynamic range (HDR) imaging has been attracting more and more attention due to its superior performance in simultaneously revealing the details in dark and bright regions. However, since most devices only support low dynamic range display, tone mapping algorithms are required to compress the dynamic range of HDR images/videos to enable watching on LDR screens \cite{ou2020real,ou2021real,ambalathankandy2019adaptive,jung2016optimized,song2018optimized}.



Traditional HDR image tone mapping methods can be classified into global tone mapping \cite{ward1994contrast,tumblin1993tone,drago2003adaptive,schlick1995quantization} and local tone mapping \cite{farbman2008edge,Gu2013,Shibata16,liang2018hybrid}. Recently, deep learning based methods have also been introduced to tone mapping. For supervised learning, the problem is how to obtain the ground truths for the input LDRs.  One solution is utilizing several available tone-mapping algorithms to generate the LDRs and the LDR result with the highest tone mapped quality index (TMQI) is selected as the ground truth \cite{cao2020adversarial}. However, the performance is limited by the upper bound of the available tone-mapping methods. Another solution regards tone mapping as low-light image enhancement task and uses the paired low-light/normal-light images for training \cite{panetta2021tmo}. Since there exists a domain gap between the low-light images and HDR images, the network trained on the enhancement dataset cannot work well for HDR tone mapping. Recently, Vinker \textit{et al.} \cite{vinker2021unpaired} propose to use unpaired HDR-LDR images for unsupervised training and has achieved promising performance. They utilize structure loss to preserve the structure consistency between HDR and LDR output, and utilize generative adversarial network (GAN) loss to force the brightness and contrast of outputs close to those of high quality LDRs. However, there still remains a lot of under-enhanced areas in their results. Therefore, developing better unsupervised training strategy is demanded. 

For HDR video tone mapping, only traditional methods are developed \cite{pattanaik2000time,boitard2012temporal,Shan2012ToneMH,kiser2012real,eilertsen2013survey,aydin2014temporally,kang2019real}. How to avoid temporal flickering and maintain rich details simultaneously is still a challenge in video tone mapping. The works in \cite{pattanaik2000time,boitard2012temporal,kiser2012real} utilize global operators which can generate results with good temporal coherency but low spatial contrast. The works \cite{benoit2009spatio,reinhard2012calibrated} using local operators can generate results with high contrast but more temporal artifacts. Therefore, developing an effective video tone mapping method to achieve a good balance between temporal consistency and richful details is demanded.





Based on the above observations, we propose a unified method for both image and video tone mapping. Our contributions are summarized as follows.

\begin{itemize}
\item{We propose an effective HDR image tone mapping network. We propose a Spatial-Feature-Enhanced (SFE) module, which utilizes graph convolution, to enable information exchange and transformation of nonlocal regions. We propose a Temporal-Feature-Replaced (TFR) module to extend our method to HDR video tone mapping which is MAC-free and can efficiently utilize the temporal correlation and improve the temporal consistency of tone-mapped results.}


\item{Unsupervised learning is difficult to be optimized and we propose a set of unsupervised loss functions to improve the result. First, we propose domain and instance based contrastive loss, in which we construct five negative and positive pairs to constrain the outputs to be close with good quality LDRs and construct a suitable latent space for measuring the similarity of the latent codes in negative and positive pairs. Second, we further propose naturalness loss to constrain the brightness and contrast of outputs.
}

\item{We construct a large-scale unpaired HDR-LDR video dataset with both real and synthetic HDR and LDR videos, which is beneficial for the development of video tone mapping. Experimental results demonstrate that our method outperforms existing state-of-the-art image and video tone-mapping methods.}
\end{itemize}

\section{Related Works}

In this section, we give a brief review of related work on HDR image and video tone mapping, image and video enhancement, and contrastive learning.

\subsection{HDR Image and Video Tone Mapping}

HDR tone mapping, which is an inverse operation of LDR to HDR reconstruction \cite{khan2019fhdr,wang2021deep}, has been widely explored in the literature. Traditional HDR image tone mapping algorithms include global tone mapping \cite{ward1994contrast,tumblin1993tone,drago2003adaptive,schlick1995quantization} and local tone mapping \cite{farbman2008edge,Gu2013,Shibata16,liang2018hybrid}. Global tone mapping \cite{ward1994contrast,tumblin1993tone,drago2003adaptive,schlick1995quantization} utilizes a global curve to compress the HDR image, which can keep the relative brightness of the input image, but usually leads to a severe reduction in local contrasts. Local tone mapping \cite{farbman2008edge,Gu2013,Shibata16,liang2018hybrid} is good at improving local contrasts and details, but usually leads to halo artifacts among high-contrast edges. For DNN-based HDR image tone mapping methods, there are mainly three categories. One category focuses on supervised learning \cite{Patel2018,Zhang19,rana2019deep,cao2020adversarial}, and they apply several tone-mapping algorithms to HDR images and select the result which has the highest TMQI \cite{yeganeh2012objective} as the ground truth. The second category regards tone mapping as image enhancement task and uses the enhancement dataset with paired data for training \cite{panetta2021tmo}. The last category gets rid of the LDR-HDR pairs, and utilizes unpaired HDR and LDR data \cite{vinker2021unpaired} or only HDR data \cite{le2021perceptually} for unsupervised training. These works either focus on the unsupervised loss functions \cite{vinker2021unpaired} or utilize no-reference image quality assessment metric to optimize the tone mapping network on HDR images \cite{le2021perceptually}.

On the basis of image TMO, traditional video tone mapping algorithms \cite{pattanaik2000time,boitard2012temporal,aydin2014temporally,Shan2012ToneMH,kiser2012real,kang2019real,boitard2014zonal} further introduce temporal processing to keep temporal stability. Although there exist lots of DNN-based HDR image TMO methods, there is still no DNN-based video TMO methods. In this work, we propose a unified unsupervised learning method for both image and video tone mapping. We further construct a large-scale unpaired HDR-LDR video dataset to facilitate the development of video tone mapping.  



\subsection{Image and Video Enhancement}
Image (video) enhancement is similar to tone mapping since it also aims at improving the brightness and contrast of the inputs.
Traditional image enhancement methods are usually histogram-based (HE)  or retinex-based methods. HE-based methods pay attention to changing the histogram of the input image to improve the brightness and contrast \cite{lee2013contrast}. Retinex-based methods decompose an input image into reflectance and illumination layers, and then enhance the image by adjusting the illumination layer \cite{guo2016lime,li2018structure,ren2020lr3m}. For DNN-based image enhancement methods, there are three kinds of training strategies which are full-supervised learning \cite{wang2019underexposed,zhang2022deep}, semi-supervised learning \cite{yang2020fidelity}, and unsupervised learning methods \cite{ignatov2018wespe,guo2020zero,li2021learning,jiang2021enlightengan}. For unsupervised learning, the work in \cite{ignatov2018wespe} maps low-quality smartphone photo to high-quality DSLR photo by utilizing CycleGAN-like architecture \cite{zhu2017unpaired}. The works in \cite{guo2020zero,li2021learning} design unsupervised loss functions to train zero-shot models. The work in \cite{jiang2021enlightengan} proposes a self-regularized perceptual loss to constrain content consistency between low-light images and enhanced images, and utilize adversarial learning to enhance contrast and brightness.
There are several DNN-based video enhancement methods \cite{lv2018mbllen,zhang2021learning}, but all of them are based on supervised learning and paired data. Since there exists distribution gap between the low-light images for image enhancement and the HDR images, directly utilizing the pretrained image (video) enhancement model for tone mapping cannot achieve satisfactory results.

\subsection{Contrastive Learning}

Contrastive learning has achieved promising progress in self-supervised and unsupervised representation learning \cite{wu2018unsupervised,he2020momentum}. It aims to improve the representation of the anchor by pushing it away from negative samples and pulling the anchor close to positive samples in the latent space. The key is how to construct positive and negative pairs, and find the latent space for distance measuring. Recently, contrastive learning has been applied to low-level vision tasks, such as image translation \cite{park2020contrastive}, super-resolution \cite{wang2021unsupervised}, dehazing \cite{wu2021contrastive}, deraining \cite{chen2022unpaired}, and underwater image restoration \cite{han2021single}.
For the strategy to construct positive and negative pairs, the work in \cite{park2020contrastive} views content consistent patches in two domains as positive pairs and views other patches as negative samples. The work in \cite{wang2021unsupervised} applies contrastive learning to unsupervised degradation estimation, taking the same and different degradations as positive and negative pairs respectively.  The works in \cite{wu2021contrastive,han2021single,chen2022unpaired} take high-quality images as positive samples, and take low-quality input images as negative samples. For the latent feature (code) for distance measuring, \cite{wu2021contrastive} utilizes VGG \cite{simonyan2014very} network to extract latent features, \cite{wang2021unsupervised} utilizes an extra encode network to code the degradation types,  \cite{park2020contrastive,wu2021contrastive,han2021single,chen2022unpaired} utilize the features extracted by the training network itself. For unsupervised image (video) tone mapping, since there are no paired supervisions, the extracted features by the generator of different images (videos) have different contents and they cannot be directly utilized for distance measuring. Therefore, we propose to aggregate the mean brightness and contrast of different channels and utilize it as the latent code of the tone-mapped images (videos).

\section{The Proposed Method}

In this section, we first introduce the network structure and then present our compound loss functions.
\begin{figure*}
    \centering
    \includegraphics[width=0.85\linewidth]{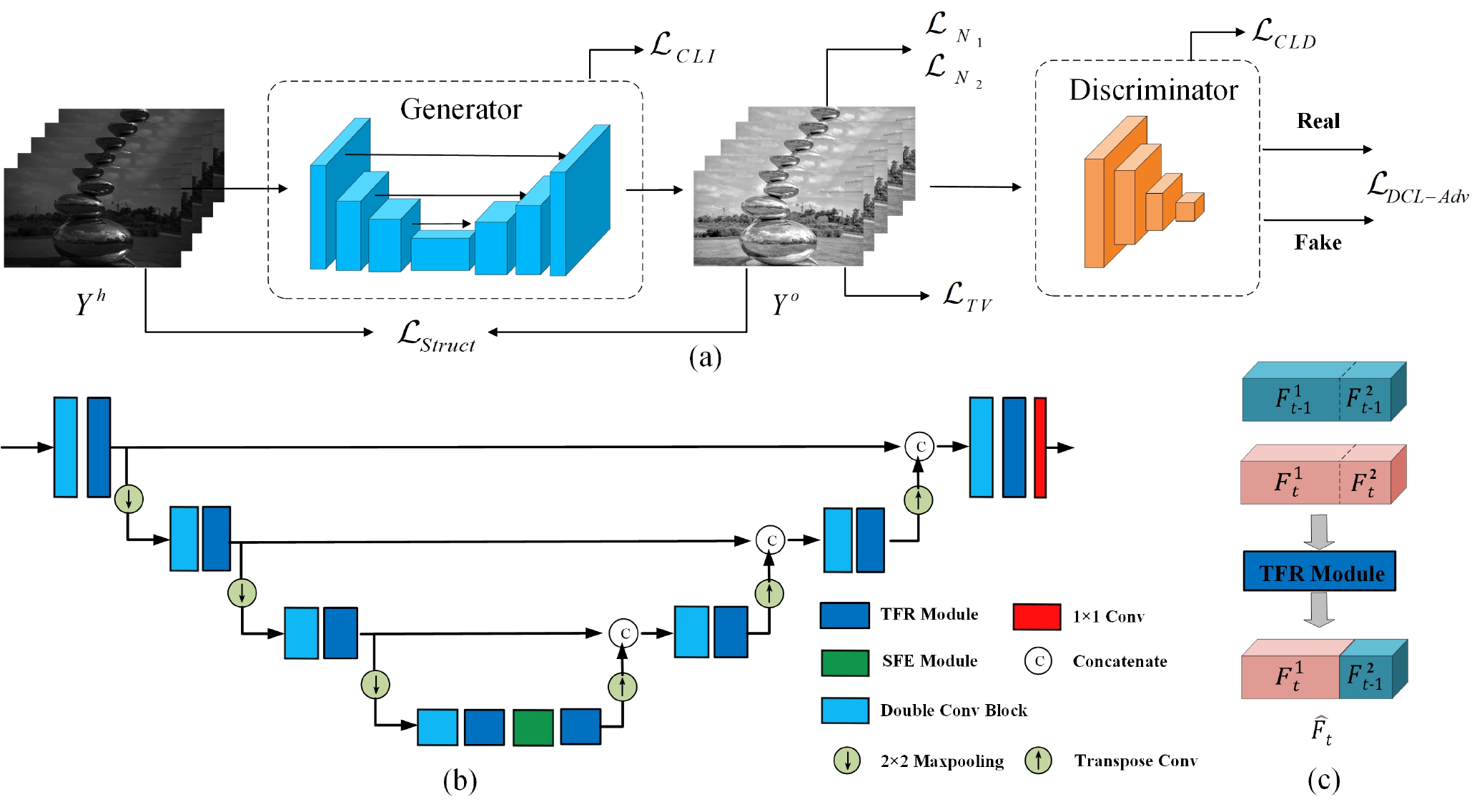}
    \caption{The proposed IVTMNet (a), which is constructed by a UNet-like generator (b) and a discriminator.  
    Note that all the operations are performed on the Y channel ($Y^h$) of the input HDR image (video) ($I^h$). The final colorful output ($I^o$) is generated by the color reproduction process. Note that, the MAC-free TFR module (c) is only utilized for video tone mapping. }
    \label{fig:framework}
\end{figure*}

\subsection{Network Structure}

Given an HDR image (video) $I^{h}$, we aim to generate its high-quality LDR image (video) $I^{o}$ through our network IVTMNet, as shown in Fig. \ref{fig:framework}. IVTMNet is constructed by a UNet-like generator and a discriminator. To capture the global statistics of brightness and contrast, we introduce the spatial feature enhanced (SFE) module at the bottom level of the UNet. For video tone mapping, we propose temporal feature replaced (TFR) module, which is beneficial for temporal consistency. It is convenient to switch between video and image TMO by removing the TFR module since this module is MAC (multiply–accumulate operation) free. 
The discriminator is utilized to distinguish the tone mapped result and it is constructed by cascaded convolutions with stride 2.
The key modules of our network are TFR and SFE, whose details are given in the following.



\subsubsection{Spatial-Feature-Enhanced (SFE) Module}
A large receptive field is beneficial for image (video) enhancement since it can capture global statistics of brightness and contrast \cite{chen2019seeing}. In this work, we adopt graph convolution proposed in \cite{han2022vision} to further enhance the spatial features and enlarge the receptive field of the network. We split the feature map into a number of patches and each patch is treated as a node. These nodes construct a graph by connecting neighboring nodes. This graph can transform and exchange information from all the nodes. In this way, similar patches with long distances can exchange and share information, which can improve the performance. Following  \cite{han2022vision}, we also apply multi-layer perceptron (MLP) module after the graph convolution for node feature transformation. To save computation cost, we only utilize the SFE module at the bottom scale of the UNet.

\subsubsection{Temporal-Feature-Replaced (TFR) Module}
Compared with image tone mapping, the major challenge for video tone mapping is making the results temporal consistent. Traditional video tone mapping methods usually solve this problem in two ways, i.e.,  utilizing flow-guided local filtering or
temporal smoothness after tone mapping. A simple temporal smoothness strategy is blending the results of previous frames with current frame
$I_t$, denoted by
\begin{equation}
\tilde{\Phi}({I_t}) = \alpha \Phi({I_t})+ (1-\alpha)\tilde{\Phi}({I_{t-1}}),
\end{equation}
where $\Phi()$ represents the tone mapping operation and $\tilde{\Phi}({I_t})$ represents the temporal filtering result of $\Phi({I_t})$. However, this may introduce ghosting artifacts. In this work, we propose a TFR module to imitate it. Recently, alignment-free module has been widely used in video denoising \cite{tassano2020fastdvdnet,qi2022real,li2022no}. TFR module can be regarded as a kind of alignment-free module. 



Specifically, for each frame $Y_{t}$, we split its corresponding feature $F_{t}$ into $F^{1}_{t}$ and $F^{2}_{t}$ along the channel dimension, as shown in Fig. \ref{fig:framework} (c). Suppose the channel number of $F_{t}$ is $q$, the channel numbers of $F^{1}_{t}$ and $F^{2}_{t}$ are $(1-\beta)q$ and $\beta q$, respectively, where $\beta$ is a splitting ratio and is set to 1/32 in our experiments. Then, we replace $F^{2}_{t}$ by $F^{2}_{t-1}$, namely that we concatenate $F^{1}_{t}$ and $F^{2}_{t-1}$ constructing the temporal enhanced feature $\hat{F}_t$ for the $t^{\text{th}}$ frame.

\begin{equation}
\begin{split}
&F^{1}_{t}, F^{2}_{t} = \text{Split}(F_{t}, \beta)\\
&F^{1}_{t-1}, F^{2}_{t-1} = \text{Split}(F_{t-1},\beta)\\
&\hat{F}_{t} = \text{Concat}(F^{1}_{t},F^{2}_{t-1})\\
\end{split}
\label{eq:tfr}
\end{equation} 
As Fig. \ref{fig:framework} shows, we insert the TFR module after each feature extraction block at different scales. The following convolution filters can selectively utilize the feature $F^{2}_{t-1}$ to help reduce the flicking artifacts and avoid ghosting artifacts. For example, for the completely static regions, this operation can help reduce noise since two frames with the same objects but different noise are fused via the following filters. Since the parameter $\beta$ (which is set to 1/32) is relatively small and we use structure loss to constrain the results to be close with the current frame, the ghosting artifacts, which may be introduced for fast motion areas, can be removed by following filters.
Therefore, the proposed TFR can help reduce the flicking artifacts and avoid ghosting artifacts. Note that, during testing, we utilize a buffer to save the features of the previous frame, which can be recurrently used for predicting the result of the next frame.

\subsubsection{Color Reproduction}

Inspired by \cite{fattal2002gradient,schlick1994quantization,tumblin1999lcis,vinker2021unpaired}, we perform tone-mapping on the luminance channel (Y) in YUV space. We denote the input and output Y channel as $Y^{h}$ and $Y^{o}$, respectively. $Y^{o}$ is converted to $I^{o}$ by a color reproduction process\cite{fattal2002gradient,schlick1994quantization,tumblin1999lcis,vinker2021unpaired}. Specifically, $I_{i}^{o}=(I_{i}^{h}/Y^{h})^{\nu} Y_{o}$, where $i$ denotes R, G, and B channel index, respectively and $I^{h}$ is the original HDR input. $\nu$ is the color saturation parameter, which is set to 0.5. In this way, the color of the original HDR can be preserved.


\begin{figure*}
    \centering
    \includegraphics[width=1.0\linewidth]%
{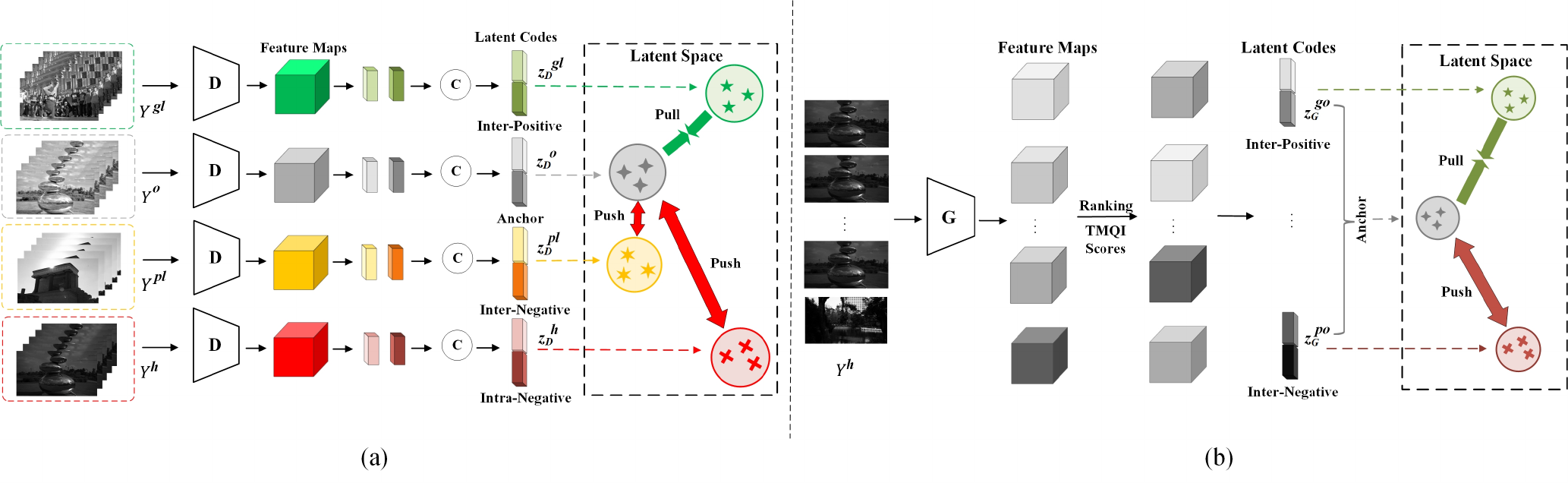}
    \caption{The proposed domain based contrastive learning (a) and instance based contrastive learning (b). The ranking index in (b) is generated based on the TMQI scores of the tone mapped results by $G$ and then their corresponding feature maps are ranked according to the index. For image TMO, the video inputs are replaced with image inputs.}
    \label{fig:contrastiveloss}
\end{figure*}

\subsection{Loss Functions}
Since there are no perfect HDR-LDR image (video) pairs for supervised learning, we propose unsupervised losses to optimize the network. Specifically, five losses are involved, including structure loss, adversarial loss, contrastive learning loss, naturalness loss, and total variation loss. The following are the details for the five loss functions.


\subsubsection{Structure Loss}
The tone mapping process should only change the brightness of the objects and not change the content. 
Therefore, we adopt the structure loss proposed in \cite{vinker2021unpaired} to preserve the content and structure between the input HDR data and output of the network. As mentioned in \cite{vinker2021unpaired}, since the SSIM metric \cite{wang2004image} is not invariant to changes in brightness and contrast, it is not suitable for the tone-mapping task, where the output is expected to undergo major changes in both brightness and contrast. Hence, we use Pearson correlation \cite{schober2018correlation} rather than SSIM to measure the structural similarity between two images, which can be formulated as
\begin{equation}
\label{eq:struct_local}
\rho(I_{1},I_{2}) = \frac{1}{n_p} \sum_{p_{I_1},p_{I_2}} \frac{\mathrm{cov}(p_{I_1}, p_{I_2})}{\sigma(p_{I_1})\sigma(p_{I_2})},
\end{equation}
where $p_{I_1}$ and $p_{I_2}$ are patches in images $I_1$ and $I_2$, $\text{cov($\cdot$,$\cdot$)}$ and $\sigma(\cdot)$ are covariance and standard deviation computed on patches, $n_p$ is the total number of patches.  
In our implementation, the patch size is 5$\times$5 and the step size for patch extraction is 1.
It is extended to structure loss as:
\begin{equation}
\label{eq:struct}
\mathcal{L}_{Struct} = \sum_{t=1}^T\sum_{k\in\{0,1,2\}}\rho(\downarrow^{k}Y^{h}_{t},\downarrow^{k}Y^{o}_{t}),
\end{equation}
where $\downarrow^{k}$ represents downsampling operations for the  $k^{\text{th}}$ spatial scale (k=0 represents the original resolution and $k$=1 represents the 1/2 downsampled resolution) and $t$ represents the temporal index for consecutive frames in video. For image tone mapping, $T$ is 1.

\subsubsection{Adversarial Loss}
We utilize adversarial learning to help the network generate pleasant results. 
We train our discriminator network $D$ (whose detailed network structure is presented in the supp. file) to distinguish between the generator’s output $Y^{o}$ and unpaired good quality LDR data $Y^{gl}$.
The generator $G$ is trained to generate good images (videos) to fool the discriminator. Different from \cite{vinker2021unpaired}, which uses least-squares GAN \cite{mao2017least}, we adopt the Dual Contrastive GAN \cite{yu2021dual} for training since it can further improve the feature representations.
The discriminator loss is formulated as
\begin{equation}
\begin{aligned}
\mathcal{L}_{DCL-D} &= \mathbb{E}_{Y^{gl}}\left[\log \frac{e^{D(Y^{gl})}}{e^{D(Y^{gl})}+\sum_{Y^{o}} e^{D(Y^{o})}}\right] \\
&+\mathbb{E}_{Y^{o}}\left[\log \frac{e^{-D(Y^{o})}}{e^{-D(Y^{o})}+\sum_{Y^{gl}} e^{-D(Y^{gl})}}\right],
\end{aligned}
\label{eq:discriminator}
\end{equation}
where $\sum_{x}f(x)$ represents the sum of $N$ processed samples $f(x)$. $\mathbb{E}_{Y^{o}}$ represents computing the expectation or mean value in square brackets for all $Y^{o}$.
The first item aims at teaching the discriminator to disassociate a single good quality LDR image (video) against a batch of generated images (videos). The second item aims at
disassociating a single generated image (video) against a batch of good quality LDR images (videos).
The generator loss can be correspondingly formulated as
\begin{equation}
\begin{aligned}
\mathcal{L}_{DCL-G} &= \mathbb{E}_{Y^{gl}}\left[\log \frac{e^{-D(Y^{gl})}}{e^{-D(Y^{gl})}+\sum_{Y^{o}} e^{-D(Y^{o})}}\right] \\
&+\mathbb{E}_{Y^{o}}\left[\log \frac{e^{D(Y^{o})}}{e^{D(Y^{o})}+\sum_{Y^{gl}} e^{D(Y^{gl})}}\right],
\end{aligned}
\label{eq:generator}
\end{equation}
The final adversarial loss can be formulated as
\begin{equation}
\begin{aligned}
\mathcal{L}_{DCL-Adv} = \mathcal{L}_{DCL-D} + \lambda_{Adv} \mathcal{L}_{DCL-G}
\end{aligned}
\end{equation}
where $\lambda_{Adv}$ is the weighting parameter (which is set to 0.1) to control the ratio between discriminator and generator loss.

\subsubsection{Contrastive Learning Loss}
\label{sec:CL}
In this work, we propose domain-based and instance-based contrastive learning (CL) losses for tone mapping, which is illustrated in Fig. \ref{fig:contrastiveloss}.

\bt{Domain-based Contrastive Learning Loss}
For low level vision tasks, CL methods usually assume low-quality inputs as negative samples and high-quality targets as positive samples, and make the output (anchor) to be close to the targets and far away from the inputs \cite{wu2021contrastive,chen2022unpaired,han2021single}.
However, the output could easily fall into the third domain: enough far away from the input domain, but not very close to the target domain. To avoid falling into this situation in tone mapping, besides the input HDR, we further select LDR data with poor quality (such as unpleasant brightness and contrast) from the public dataset as another negative domain. The poor LDRs can be regarded as failure cases of tone mapping. We push the output not only far away from the input HDR domain but also far away from the poor LDR domain. In other words, our negative pairs consist of two kinds, i.e., the outputs and the inputs, the outputs and the poorly-exposed LDRs. Our positive pairs are constructed by the outputs and the well-exposed LDRs.

The other question is finding the latent space for distance measuring. Considering that our discriminator has learned feature representations to distinguish the outputs of our generator and good quality LDRs, we utilize the discriminator ($D$) to extract features for CL, other than utilizing a universal VGG network \cite{simonyan2014very}\cite{wu2021contrastive}. Specifically, we utilize the features after the last convolution layer of $D$ but before being flattened into a vector. Then we calculate the mean value ($\mu_i$) and the mean contrast ($\tau_i$, which is calculated according to \cite{wang2004image}) of each channel ($C_i$) and concatenate them together to build the latent code, denoted as $z=[\mu_1,\mu_2,...\mu_q,\tau_1,\tau_2,...,\tau_q]$, where $q$ is the channel number of the convolution feature. In this way, we construct the latent codes for $Y^o$, $Y^{gl}$ (the good quality LDR data), $Y^{pl}$ (the poorly exposed LDR data), and $Y^{h}$, and they are denoted as $z_D^{o}$, $z_D^{gl}$, $z_D^{pl}$, and $z_D^{h}$ respectively, where $D$ represents these latent codes are from the discriminator features. Since $Y^o$ and $Y^h$ are different formats of the same content, $z_D^{h}$ can be viewed as the intra-negative sample of the anchor $z_D^{o}$. Similarly, $z_D^{gl}$ is the inter-positive sample and $z_D^{pl}$ is the inter-negative sample since they have different contents from the anchor. Therefore, our domain-based contrastive learning loss can be formulated as
\begin{equation}
\begin{aligned}
\mathcal{L}_{CLD} &= \mathbbm{E} \left[-\log \frac{\operatorname{s}\left(z^{o}_{D}, z^{gl}_{D}\right)}{\operatorname{s}\left(z^{o}_{D}, z^{gl}_{D}\right)+\sum_{i=1}^{N} \operatorname{s}\left(z^{o}_{D}, z^{h_i}_{D}\right)}\right] \\
&  + \mathbbm{E} \left[-\log \frac{\operatorname{s}\left(z^{o}_{D}, z^{gl}_{D}\right)}{\operatorname{s}\left(z^{o}_{D}, z^{gl}_{D}\right)+\sum_{i=1}^{N} \operatorname{s}\left(z^{o}_{D}, z^{pl_i}_{D}\right)}\right],
\end{aligned}
\label{eq:CLD}
\end{equation}
where the first (second) term utilizes $z_D^h$ ($z_D^{pl}$) as negative samples and $N$ is the negative sample number.
Inspired from the cosine distance, we propose a new cosine distance by incorporating the $\ell_1$ distance as denominator to measure the similarity between the latent codes, i.e.,
\begin{equation}
\operatorname{s}(u, v)=\exp \left(\frac{u^{T} v}{\eta+c|u-v|_1}\right),
\label{eq:sim}
\end{equation}
where $\eta$ and $c$ are hyper-parameters.
We would like to point out, as the GAN training, the capability of the discriminator to distinguish the outputs and the good quality LDRs will be weakened. However, the discriminator will still be good at discriminating the outputs and the poorly-exposed LDRs, and the outputs and the inputs. Therefore, Eq. \ref{eq:CLD} will still work by pushing the output away from the inputs and the poorly-exposed LDRs.

\bt{Instance-based Contrastive Learning Loss}
Besides contrastive learning with images (videos) from different domains, we further make the generator learn from itself dynamically. We observe that for one batch of images (videos), some samples are tone-mapped well while some are not. Therefore, we propose instance-based contrastive learning to push the output apart from the poorly tone-mapped samples and pull the output close to the good quality tone-mapped samples. We utilize TMQI \cite{yeganeh2012objective} metric to evaluate the tone-mapped result. For each batch, the image with the highest TMQI score is selected as the positive sample, and the image with the lowest TMQI score is selected as the negative sample. Since the results of one batch have different TMQI scores, their features before the final output layer of the generator also contain many differences. Therefore, we utilize the generator's feature maps before the last convolution to construct the latent codes. The construction process is the same as that in domain-based contrastive learning loss. The latent codes for the inter-positive and inter-negative samples are denoted as $z_G^{go}$ and $z_G^{po}$, respectively. The latent codes for the whole batch are denoted as $z_G^{o}$. In this way, our instance-based contrastive learning loss can be formulated as
\begin{equation}
\begin{aligned}
\mathcal{L}_{CLI} &= \mathbbm{E} \left[-\log \frac{\operatorname{s}\left(z^{o}_{G}, z^{go}_{G}\right)}{\operatorname{s}\left(z^{o}_{G}, z^{go}_{G}\right)+ \operatorname{s}\left(z^{o}_{G}, z^{po}_{G}\right)}\right],
\end{aligned}
\label{eq:CLI}
\end{equation}
where $\operatorname{s}(u, v)$ is the same as that defined in Eq. \ref{eq:sim}.

\subsubsection{Naturalness Loss}
We further propose the naturalness loss by measuring the naturalness of an image with its mean brightness and mean contrast, where the mean brightness is calculated based on the mean value (${{m}}(\cdot)$) and the contrast is calculated based on the variance ($\sigma^2(\cdot)$) (this process is similar to that used in SSIM \cite{wang2004image}). Specifically, the distances are calculated based on patches (the patch size is $11\times11$ and the patch extraction step is 1) as follows.
\begin{equation}
\label{eq:natural_contrast}
\phi_{\sigma}(I_{1},I_{2}) = \frac{1}{n_p} \sum_{p_{I_1},p_{I_2}} |\sigma^2(p_{I_1})-\sigma^2(p_{I_2})|_1,
\end{equation}
\begin{equation}
\label{eq:natural_bright}
\phi_{m}(I_{1},I_{2}) = \frac{1}{n_p} \sum_{p_{I_1},p_{I_2}} |m(p_{I_1})-m(p_{I_2})|_1,
\end{equation}
where $\phi_{\sigma}(I_{1},I_{2})$ and $\phi_{m}(I_{1},I_{2})$ denote the contrast and mean brightness differences between the two images, respectively.
On the one hand, we constrain the naturalness of the output to be similar with that of good quality LDR data (namely inter supervision) since they have pleasant brightness and contrast. This loss is formulated as
\begin{equation}
\begin{aligned}
\mathcal{L}_{N_1} = \sum_{t=1}^T\phi_{\sigma}(Y^{gl}_{t}, Y^{o}_{t})+\phi_{m}(Y^{gl}_{t}, Y^{o}_{t}).
\end{aligned}
\end{equation}
On the other hand, we observe that for one image, there exist some areas well-tone-mapped. 
Therefore, we split the output into four patches (namely 2$\times$2 patches) and the patch size depends on the image resolution. Then, we select the patch with the highest TMQI socre as the intra label, denoted as $Y^{gp}_{t}$.
This loss is formulated as
\begin{equation}
\begin{aligned}
\mathcal{L}_{N_2} = \sum_{t=1}^T\phi_{\sigma}(Y^{gp}_{t},Y^{o}_{t})+\phi_{m}(Y^{gp}_{t},Y^{o}_{t})
\end{aligned}
\end{equation}
The two terms together form our naturalness loss. For image TMO, the $T$ in the above two formulas is equal to 1. We utilize TMQI score, other than other measurements (such as NIQE) is because the TMQI is specifically designed for HDR TMO assessment and can better rank the outputs.

\subsubsection{Total Variation Loss}
To remove magnified noise in dark regions, we further apply total variation (TV) loss to the generator's output $Y^{o}_{t}$, denoted as
\begin{equation}
\label{equ_7}
\mathcal{L}_{TV} = \sum_{t=1}^T (|\nabla_{x}Y^{o}_{t}|+|\nabla_{y}Y^{o}_{t}|)^2,
\end{equation}
where $T$ is the number of video frames in video TMO, and is equal to 1 in image TMO. $\nabla_{x}$ and $\nabla_{y}$ represent the horizontal and vertical gradient operations, respectively.

With the aforementioned loss functions, our full loss function can be denoted as
\begin{equation}
\begin{aligned}
\mathcal{L} &= \mathcal{L}_{Struct} + \lambda_1 \cdot \mathcal{L}_{DCL-Adv} + \lambda_2 \cdot \mathcal{L}_{CLD} + \lambda_3 \cdot \mathcal{L}_{CLI} \\
& + \lambda_4 \cdot \mathcal{L}_{N_1} + \lambda_5 \cdot \mathcal{L}_{N_2} + \lambda_6 \cdot \mathcal{L}_{TV}
\end{aligned}
\end{equation}
where the weighting parameters $(\lambda_1 ... \lambda_6)$ are used to control the ratio of each loss.

\section{Dataset}

\subsection{Image Tone Mapping Dataset}
Following \cite{vinker2021unpaired}, we also utilize 1000 HDR images from HDR+ dataset and the well-exposed LDR images from DIV2K dataset \cite{agustsson2017ntire} for image tone mapping training. Since our domain-based CL loss needs poorly-exposed LDR data, we further utilize 1300 images with unpleasant brightness and contrast from \cite{cai2018learning} as poorly-exposed LDR images for training. Follow the settings of \cite{vinker2021unpaired}, we crop and rescale every training image to a size of 256$\times$256. We evaluate the performance on the HDR Survey \cite{HDRsurvey}, HDRI Haven \cite{zaal} and LVZ-HDR dataset \cite{panetta2021tmo}, which is the same as \cite{vinker2021unpaired}.

\subsection{Video Tone Mapping Dataset}
\label{sec:videodataset}
There is no available supervised or unsupervised dataset for video tone mapping and there is no perfect solution to generate ground truth LDR videos for HDR inputs. Therefore, we construct the unpaired HDR-LDR video dataset to enable the unsupervised training of our network.

First, we construct the unpaired HDR-LDR video dataset with real captured videos. We collect HDR videos from traditional HDR tone mapping works \cite{froehlich2014creating,eilertsen2016evaluation} and HDR reconstruction works \cite{kalantari2013patch,kronander2013unified,chen2021hdr}. Considering that the HDR videos with large flicker artifacts will affect the evaluation of tone mapping algorithms, we remove them and totally collect 100 HDR videos with high quality. For LDR videos, we select 80 well-exposed LDR videos from DAVIS dataset \cite{perazzi2016benchmark} for adversarial learning and constructing positive pairs in contrastive learning. Most of our collected HDR videos are 1280$\times$720, and we resize the other videos (1920$\times$1080 \cite{froehlich2014creating}, 1476$\times$753 \cite{chen2021hdr}) to 1280$\times$720 to unify the resolution, which will simplify the following patch cropping process. The 100 HDR videos are split into training and validation set (80 videos), and testing set (20 videos).


Considering that both adversarial learning and contrastive learning require a large amount of training data to improve the optimization process, we further construct synthesized HDR and LDR videos. Specifically, we synthesize videos from static images by dynamic random cropping. We utilize 1000 HDR images from HDR+ dataset \cite{hasinoff2016burst}, 780 good quality LDR images from DIV2K dataset \cite{agustsson2017ntire} and 1300 poorly LDR images with unpleasant brightness and contrast from \cite{cai2018learning}. For one image, we first randomly downsample it with a ratio of $\gamma$ and then randomly crop $T$ patches with a resolution of $256\times256$ to construct a sequence with $T$ frames. The downsampling ratio $\gamma$ ranges from 1 to 2.8. When $\gamma$ is small, the probability that the cropped patches have overlapped regions is small. In this case, we simulate videos with large movements and vice versa. The synthesized poorly-LDR videos are used for negative pair construction in CL. The good quality LDR videos are used for adversarial learning and positive pair construction.

In summary, our video tone mapping dataset contains 1100 HDR videos, 860 good quality LDR videos, and 1300 poorly LDR videos. 
Among them, 20 real captured HDR videos are selected for testing. During testing, we directly process the video with its original resolution.




\section{Experiments}

\subsection{Training Details}
During image TMO training, each image is cropped and rescaled to provide two 256$\times$256 images, which is the same as \cite{vinker2021unpaired}. During video TMO training, we resize the real HDR and LDR videos from 1920$\times$1080 to 455$\times$256, and randomly crop 256$\times$256 sequences for training. The synthetic HDR and LDR videos are also downsampled and then cropped to 256$\times$256 squences, which has been described in Section \ref{sec:videodataset}. Therefore, the training patches has a large dynamic range.
The batch size is set to 8. 
For video tone mapping, the frame number $T$ is set to 3, which can ensure temporal stability and avoid ghosting artifacts.
We train our generator and discriminator with learning rate 1e-5 and 1.5e-5 respectively, and they decay by half for every 10 epochs. For each iteration in the training process, we first optimize the discriminator by maximizing Eq. \ref{eq:discriminator} and then optimize the generator by minimizing Eq. \ref{eq:generator}.
The sample number $N$ in Eqs. \ref{eq:discriminator}, \ref{eq:generator} and \ref{eq:CLD} is set to 16.
The parameters $\eta$ and $c$ in Eq. \ref{eq:sim} are set to 1e-2 and 1. The weighting parameters $\lambda_1 ... \lambda_6$ for different loss functions are dynamically updated. At the early stage (smaller than 7 epochs), we optimize the network mainly by adversarial learning and contrastive learning. Therefore, $\lambda_1 ... \lambda_6$ are set to 1, 0.5, 0.1, 0.001, 0.001, 0.001, respectively. At the middle stage (7-10 epochs), we magnify the weight $\lambda_4$ (by setting it to 0.5), which forces the network to generate better brightness and contrast. At the later stage (10-20 epochs), we further force the network to learn from itself and $\lambda_5, \lambda_6$ are set to 0.5 and 0.2, respectively. Note that, since the second naturalness loss depends on the output, $\lambda_5$ is magnified later than $\lambda_4$. We would like to point out that our final results are not sensitive to the change of weighting parameters. If we fix $\lambda_1 ... \lambda_6$ for the whole time, we can still get good results, which is only slightly worse than our current setting.

\subsection{Ablation Study}

\begin{table}
\centering
\caption{Ablation study for proposed loss functions, TFR module and SFE module.}
\resizebox{0.49\textwidth}{16mm}{
\begin{tabular}{ccccccccccc}
\toprule
\multicolumn{2}{l}{$\mathcal{L}_{Struct}$}  &$\checkmark$ &$\checkmark$ &$\checkmark$ &$\checkmark$ &$\checkmark$ &$\checkmark$ &$\checkmark$ &$\checkmark$ &$\checkmark$\\\hline
\multicolumn{2}{l}{$\mathcal{L}_{DCL-Adv}$} &$\times$     &$\checkmark$ &$\checkmark$ &$\checkmark$ &$\checkmark$ &$\checkmark$ &$\checkmark$ &$\checkmark$ &$\checkmark$\\\hline
\multirow{2}{*}{$\mathcal{L}_{CL}$} &$\mathcal{L}_{CLD}$     &$\times$     &$\times$     &$\checkmark$ &$\checkmark$ &$\checkmark$ &$\checkmark$ &$\checkmark$ &$\checkmark$ &$\checkmark$\\ \cline{2-11}
                                    &$\mathcal{L}_{CLI}$     &$\times$     &$\times$     &$\times$     &$\checkmark$ &$\checkmark$ &$\checkmark$ &$\checkmark$ &$\checkmark$ &$\checkmark$\\\hline
\multirow{2}{*}{$\mathcal{L}_{N}$}  &$\mathcal{L}_{N_1}$     &$\times$     &$\times$     &$\times$     &$\times$     &$\checkmark$ &$\checkmark$ &$\checkmark$ &$\checkmark$ &$\checkmark$\\ \cline{2-11}
                                    &$\mathcal{L}_{N_2}$     &$\times$     &$\times$     &$\times$     &$\times$     &$\times$     &$\checkmark$ &$\checkmark$ &$\checkmark$ &$\checkmark$\\\hline
\multicolumn{2}{l}{$\mathcal{L}_{TV}$}      &$\times$     &$\times$     &$\times$     &$\times$     &$\times$     &$\times$     &$\checkmark$ &$\checkmark$ &$\checkmark$\\\hline
\multicolumn{2}{l}{TFR module}                  &$\checkmark$ &$\checkmark$ &$\checkmark$ &$\checkmark$ &$\checkmark$ &$\checkmark$ &$\checkmark$ &$\times$     &$\times$    \\\hline
\multicolumn{2}{l}{SFE module}                   &$\checkmark$ &$\checkmark$ &$\checkmark$ &$\checkmark$ &$\checkmark$ &$\checkmark$ &$\checkmark$ &$\checkmark$ &$\times$    \\\hline
\multicolumn{2}{l}{TMQI$\uparrow$}                    & 0.8939 & 0.8970 & 0.8995 & 0.9017 & 0.9025 & 0.9036  & 0.9052 & 0.9038 & 0.9027 \\
\multicolumn{2}{l}{RWE($10^{-2}$)$\downarrow$}        & 4.72   & 4.52   & 5.04   & 4.98   & 4.82   & 4.66    & 4.38   & 4.92   & 5.10   \\
\bottomrule
\end{tabular}
}
\label{Ablation}
\end{table}

\begin{table}[t]
\centering
\caption{Ablation study for the detailed settings of our CL loss and naturalness loss.}
\resizebox{0.45\textwidth}{21mm}{
\begin{tabular}{clccc}
\toprule
\multirow{3}{*}{$\mathcal{L}_{CLD}$}& Poorly-LDR negative samples     & $\times$ & $\checkmark$  \\
\multirow{3}{*}{}                    & TMQI$\uparrow$                 & 0.8984  & 0.8995  \\
\multirow{3}{*}{}                    & RWE($10^{-2}$)$\downarrow$     & 5.30    & 5.04  \\
\midrule
\multirow{3}{*}{$\mathcal{L}_{CL}$} & Our distance                   & $\checkmark$ & $\times$     & $\checkmark$\\
\multirow{3}{*}{}                   & Our latent code                & $\times$     & $\checkmark$ & $\checkmark$\\
\multirow{3}{*}{}                   & TMQI$\uparrow$                 & 0.8996 & 0.8981 & 0.9017 \\
\multirow{3}{*}{}                   & RWE($10^{-2}$)$\downarrow$     & 5.22   & 4.88 & 4.98 \\
\midrule
\multirow{3}{*}{$\mathcal{L}_{N_1}$}& Mean Brightness                & $\times$ & $\checkmark$ & $\checkmark$ \\
\multirow{3}{*}{}                   & Mean Contrast                  & $\times$ & $\times$     & $\checkmark$ \\
\multirow{3}{*}{}                   & TMQI$\uparrow$                 & 0.9017   & 0.9022  & 0.9025  \\
\multirow{3}{*}{}                   & RWE($10^{-2}$)$\downarrow$     & 4.98     & 4.83    & 4.82  \\
\bottomrule
\end{tabular}
}
\label{tab:moreAblation}
\end{table}

In this section, we perform ablation study to demonstrate the effectiveness of the proposed loss functions and TFR, SFE modules. To explore the influence of each component for the tone mapping and temporal consistence, we perform ablation study on our video TMO test data. We utilize two measurements to evaluate the tone-mapping quality. The first is the widely used TMQI \cite{yeganeh2012objective} score. The second is warping error (WE) \cite{lai2018learning}, which is widely used to evaluate the temporal consistence of processed frames. However, it tends to reward the tone-mapping methods that generate dark results which are not the aim of the tone-mapping task. Therefore, we propose the relative warping error (RWE) measurement by dividing the warping error by the mean intensity, which is formulated as
\begin{equation}
\begin{aligned}
\bar{Y}^o_{t} &= \psi({Y}^o_{t},{Y}^o_{t-1}) \\
\text{RWE} &= \frac{2}{HW} \sum_{i=1}^H \sum_{j=1}^W \frac{|{Y}^o_{t-1}(i,j)-\bar{Y}^o_{t}(i,j)|}{{Y}^o_{t-1}(i,j)+\bar{Y}^o_{t}(i,j)},
\end{aligned}
\end{equation}
where $\psi(x,y)$ denotes warping x towards y, and $\bar{Y}^o_{t}$ is the warping result. $H$ and $W$ denote the height and width of ${Y}^o_{t}$. In this work, we utilize deepflow \cite{weinzaepfel2013deepflow} to serve as the warping operation.  For each video, we select the first six frames and calculate the mean TMQI and RWE scores for all the testing videos. The results are given in  Table \ref{Ablation}. For the  baseline model ($\mathcal{L}_{Struct}$), we utilize structure consistency loss and GAN loss. It can be observed that the TMQI score is consistently improved by introducing the proposed loss functions. After introducing the proposed contrastive loss, the RWE score is increased since more details are revealed. Fortunatelly, after introducing the TV loss, the RWE score is reduced and TMQI score is also increased slightly. When the TFR module is removed , the TMQI score is degraded and the RWE value is increased from 4.38 to 4.92. It verifies that the proposed TFR module is beneficial for temporal consistency. In addition, the proposed SFE module can improve the TMQI score since it encourages information exchange in non-local regions.

\begin{figure}
    \centering
    \includegraphics[width=1.0\linewidth]{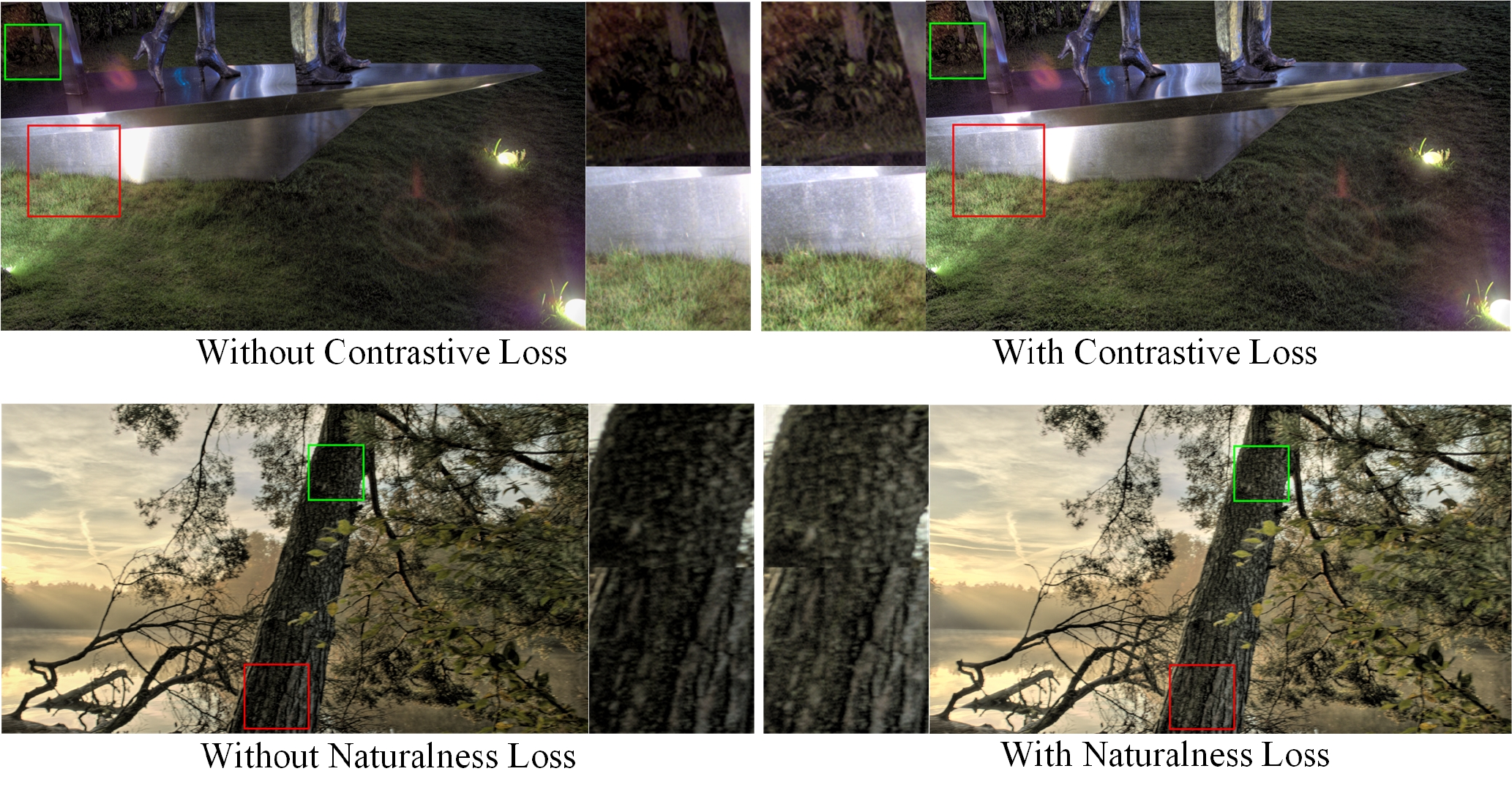}
    \caption{Visual quality comparison for the ablation of the proposed contrastive learning loss and naturalness loss.}
    \label{fig:losscompare}
\end{figure}

We further give the ablation for the detailed settings of our contrastive learning (CL) loss and naturalness loss.

1) \textbf{Negative pair ablation.} CL methods for low-level vision usually assume low-quality inputs as negative samples and high-quality targets as positive sample \cite{wu2021contrastive,chen2022unpaired,han2021single}. Different from them, we utilize extra poorly-exposed LDRs as negative samples to calculate our domain-based contrastive learning loss, which has been formulated as Eq. \ref{eq:CLD}. The first and the second term utilizes input HDR $z_D^h$ and poorly-exposed LDRs $z_D^{pl}$ as negative samples, respectively. To analyze the influence of poorly-LDR negative samples, we remove the second term, and the results are given in the first row of Table \ref{tab:moreAblation}. When the poorly-LDR negative samples are removed, the TMQI score is decreased and the RWE value is increased, which demonstrates that the introduced poorly-LDR samples are important to the success of contrastive learning.

2) \textbf{Latent code ablation.} In our CL loss, our latent code is built by the the mean value ($\mu_i$) and the mean contrast ($\tau_i$) of each feature channel ($C_i$). Here, we evaluate it by replacing it by an intuitive setting. Specifically, we replace our latent code with global averaging of feature maps. It can be observed that the TMQI score is degraded by replacing our latent code with the intuitive one, as shown in Table \ref{tab:moreAblation}.

3) \textbf{Distance measurement ablation.} We propose a new distance measurement to calculate the similarity between latent codes, which has been formulated as Eq. \ref{eq:sim}. 
To demonstrate the effectiveness of the distance measurement, we replace it with cosine distance. As shown in Table \ref{tab:moreAblation} (the second column of $\mathcal{L}_{CL}$ ablation), our proposed distance measurement is much better than cosine distance. In the supp. file, we further give the ablation results by replacing our distance measurement by the popular $\ell_1$ and $\ell_2$ distance measurements. Our result is still better than them.

Fig. \ref{fig:losscompare} presents visual comparison results for the ablation study. It can be observed that introducing contrastive loss can improve the contrast for both dark and over-exposed areas. 
This is mainly because our domain-based contrastive learning can push the output away from poor-quality LDR videos that contain many under-exposed and over-exposed regions.
Our instance-based contrastive learning can further make the network learn from itself to generate better results.
By introducing the naturalness loss, the dark areas can be further lightened.

\subsection{Comparison for HDR Image Tone Mapping}

\begin{table}[t]
\centering
\caption{Quantitative comparison with state-of-the-art image TMO methods on the HDR Survey dataset~\cite{HDRsurvey}. For the methods marked by $*$, we directly quote their scores from their papers. The result of \cite{le2021perceptually} is generated with the model retrained by ourselves. For the other compared methods, their scores are quoted from \cite{vinker2021unpaired}.
}
\label{tab:compareImageTMO1}
\begin{tabular}{lcc}
    \hline
    TMO & TMQI$\uparrow$ & BTMQI$\downarrow$\\
    \hline
    Ferradans \etal~\cite{ferradans2011analysis} & 0.836 & 4.563\\
    Mai \etal~\cite{mai2010optimizing} & 0.856 & 3.958\\
    Shibata \etal~\cite{Shibata16} & 0.87 & 3.578 \\
    Gu \etal~\cite{Gu2013} & 0.871 & 3.878\\
    Shan \etal~\cite{shan2009globally} & 0.874 & 3.625 \\
    Zhang \etal~\cite{Retina_inspired_2020}* & 0.88 & 3.76 \\
    Farbman \etal~\cite{farbman2008edge}& 0.886 & 3.602\\
    Liang \etal~\cite{liang2018hybrid} & 0.887 & 3.691 \\
    Khan \etal~\cite{Khan20}* & 0.889 & -- \\
    Ma \etal~\cite{TMQI_optim} & 0.895 & 3.868 \\
    Cao \etal~\cite{cao2020adversarial}* & 0.9 & -- \\
    Paris \etal~\cite{Paris11} & 0.906 & 2.988\\
    DeepTMO \cite{rana2019deep}* & 0.88 & --\\
    TMOCAN\cite{le2021perceptually} & 0.899  & 2.94  \\
    UnpairedTMO\cite{vinker2021unpaired}* & 0.919 & 2.89 \\
    \hline
    \textbf{Ours} & \textbf{0.932} & \textbf{2.83} \\
    \hline
\end{tabular}
\end{table}

\begin{table}[t]
\begin{center}
\caption{Quantitative comparison with state-of-the-art image TMO methods on the HDRI Haven dataset~\cite{zaal}. The results of \cite{farbman2008edge} and \cite{liang2018hybrid} are generated with their released codes. The results of \cite{vinker2021unpaired} and \cite{Zhang19} are quoted from \cite{vinker2021unpaired}. The result of \cite{le2021perceptually} is generated with the model retrained by ourselves.
}
\label{tab:compareImageTMO2}
\begin{tabular}{lcc}
    \hline
    TMO & TMQI$\uparrow$ & BTMQI$\downarrow$\\
    \hline
    Farbman \etal~\cite{farbman2008edge} & 0.857  & 4.58 \\
    Liang \etal~\cite{liang2018hybrid} & 0.798 & 4.94  \\
    Zhang \etal~\cite{Zhang19} & 0.874 & 3.519 \\
    TMOCAN\cite{le2021perceptually} & 0.796 & 4.16  \\
    UnpairedTMO\cite{vinker2021unpaired} & 0.902 & 3.24 \\
    \hline
    Ours & \textbf{0.916} & \textbf{3.13} \\
    \hline
\end{tabular}
\end{center}
\end{table}

\begin{table}[t]
\begin{center}
\caption{Quantitative comparison with state-of-the-art image TMO methods on the LVZ-HDR dataset~\cite{panetta2021tmo}.
The results of \cite{farbman2008edge} and \cite{liang2018hybrid} are generated with their released codes. The results of \cite{vinker2021unpaired} and \cite{panetta2021tmo} are quoted from \cite{vinker2021unpaired}. The result of \cite{le2021perceptually} is generated with the model retrained by ourselves.
}
\label{tab:compareImageTMO3}
\begin{tabular}{lcc}
    \hline
    TMO & TMQI$\uparrow$ & BTMQI$\downarrow$\\
    \hline
    Farbman \etal~\cite{farbman2008edge} & 0.822  & 5.15 \\
    Liang \etal~\cite{liang2018hybrid} & 0.881 & 3.45 \\
    TMONet \cite{panetta2021tmo} & 0.873 & 3.44 \\
    TMOCAN\cite{le2021perceptually} & 0.877  & 3.32  \\
    UnpairedTMO\cite{vinker2021unpaired} & 0.883 & 3.041 \\
    \hline
    Ours & \textbf{0.899} & \textbf{2.91} \\
    \hline
\end{tabular}
\end{center}
\end{table}

To demonstrate the effectiveness of the proposed TMO for image tone mapping, we compare with state-of-the-art image tone mapping methods on HDR Survey dataset \cite{HDRsurvey}, HDRI Haven dataset \cite{zaal} and LVZ-HDR dataset \cite{panetta2021tmo}, respectively. For the compared methods, \cite{cao2020adversarial,Zhang19,rana2019deep,panetta2021tmo} are full-reference based TMO methods, \cite{vinker2021unpaired} are unpaired TMO methods, \cite{le2021perceptually} are non-reference (zero-reference) based TMO methods. Since TMOCAN\cite{le2021perceptually} did not release the pretrained model, we retrain it on the HDR+ dataset for fair comparison. The results are shown in Tables \ref{tab:compareImageTMO1}, \ref{tab:compareImageTMO2}, and \ref{tab:compareImageTMO3}, respectively. Besides TMQI, we further utilize blind TMQI (BTMQI) \cite{gu2016blind} to evaluate the tone mapping quality. It can be observed that our method achieves the best TMQI and BTMQI scores in all three image TMO datasets. For HDR Survey dataset, our method outperforms the second best method UnpairedTMO by 0.013 and 0.06 gain for TMQI and BTMQI scores. For HDR Haven dataset, our method outperforms the second best method by 0.014 and 0.11 gain for TMQI and BTMQI scores. For LVZ-HDR dataset, our method outperforms the second best method by 0.016 and 0.131 gain for TMQI and BTMQI scores.

\begin{figure*}
    \centering
    \includegraphics[width=0.99\linewidth]{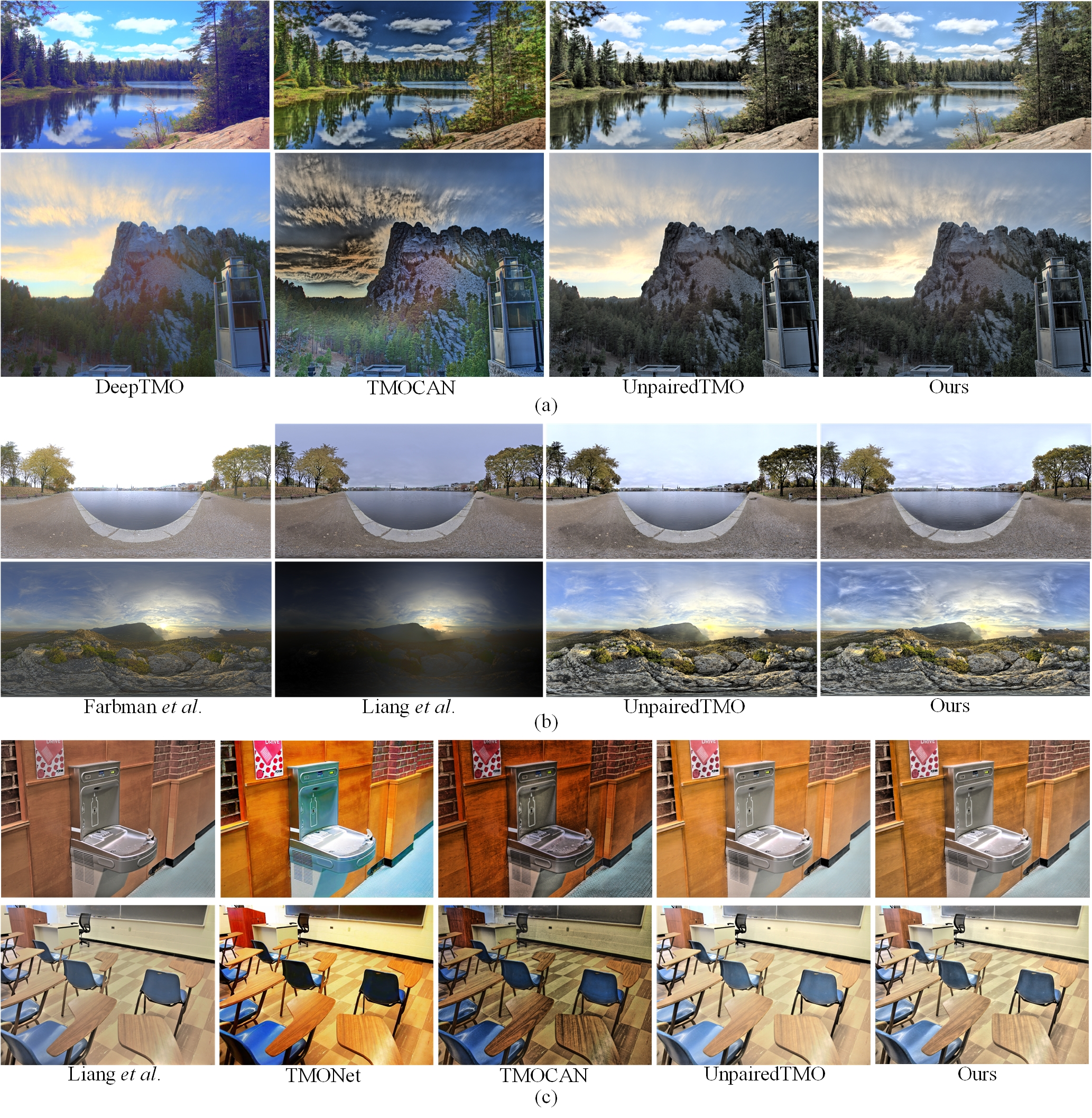}
    \caption{Visual quality comparison for HDR image tone mapping. We compare with state-of-the-art methods on the images from three datasets: (a) HDR Survey dataset, (b) HDRI Haven dataset, and (c) LVZ-HDR dataset.}
    \label{fig:methodcompareimage}
\end{figure*}

Fig. \ref{fig:methodcompareimage} present the visual comparison results on the three image TMO datasets respectively. Among them, DeepTMO \cite{rana2019deep} sometimes generates results with false colors, such as the bluish tree in the first row of Fig. \ref{fig:methodcompareimage}. The work in \cite{le2021perceptually} generates over-enhanced results which have unnatural contrast and textures. The works in \cite{farbman2008edge} and \cite{liang2018hybrid} sometimes generate under-exposed results, especially on several scenes in the HDRI Haven dataset. The work in \cite{panetta2021tmo} usually generates over-exposed results, as shown in Fig. \ref{fig:methodcompareimage} (c). UnpairedTMO \cite{vinker2021unpaired} can generate better results, but there usually exists under-exposed and over-exposed areas. Compared with these methods, our method has the best visual quality with suitable brightness and contrast.

\subsection{Comparison for HDR Video Tone Mapping}


\begin{table*}[t]
\centering
\caption{Quantitative comparison with state-of-the-art methods for video tone mapping. The best results are highlighted in bold and the second best results are underlined. All the deep learning based methods are retrained on our video HDR dataset.}
\begin{tabular}{m{3.0cm}m{2.0cm}<{\centering}m{1.5cm}<{\centering}m{1.2cm}<{\centering}m{1.2cm}<{\centering}m{1.6cm}<{\centering}m{1.6cm}<{\centering}m{2.0cm}<{\centering}}
\toprule
Methods                                    & Supervision & Image/Video & TMQI$\uparrow$  & BTMQI$\downarrow$ & RWE($10^{-2}$)$\downarrow$ & MACs($10^{9}$)$\downarrow$ & Parameters (M)$\downarrow$ \\
\hline
Farbman \etal~\cite{farbman2008edge}                     & Traditional &Image & 0.8701 & 4.4283   & 4.74 & - & -\\
Liang \etal~\cite{liang2018hybrid}                     & Traditional &Image & 0.8429 & 3.3534   & 4.52 & - & -\\
Kiser \etal~\cite{kiser2012real}                       & Traditional &Video & 0.7901 & 4.8242   & \textbf{3.24} & - & -\\
TMOCAN \cite{le2021perceptually}           & Zero-Reference &Image & 0.8611 & $3.1364$   & 5.64 & 3.27 & \textbf{0.074}\\
UnpairedTMO \cite{vinker2021unpaired}      & Unpaired &Image & $0.8922$ & 3.2834   & 5.70 & 10.06 & 4.489\\
\hline
ZeroDCE \cite{guo2020zero}                 & Zero-Reference &Image & 0.7789 & 4.9504   & 6.00 & 5.19 & $\underline{0.079}$\\
EnlightenGAN \cite{jiang2021enlightengan}  & Unpaired &Image & 0.7510 & 4.8904   & 5.20 & 16.45 & 8.637\\
\hline
Ours (IVTMNet$_{\text{full}}$)                     & Unpaired &Video & \textbf{0.9052} & \textbf{2.9978}        & $\underline{4.38}$ & $10.11$ & 4.884\\
Ours (IVTMNet$_{0.75}$)                     & Unpaired &Video & $\underline{0.8991}$ & $\underline{3.0296}$   & 4.40               & $\underline{5.69}$ & 2.748\\
Ours (IVTMNet$_{0.5}$)                      & Unpaired &Video & 0.8972  & 3.0479                 & 4.61               & \textbf{2.53} & 1.222\\
\bottomrule
\end{tabular}
\label{Comparison}
\end{table*}

\begin{figure*}
    \centering
    \includegraphics[width=0.99\linewidth]{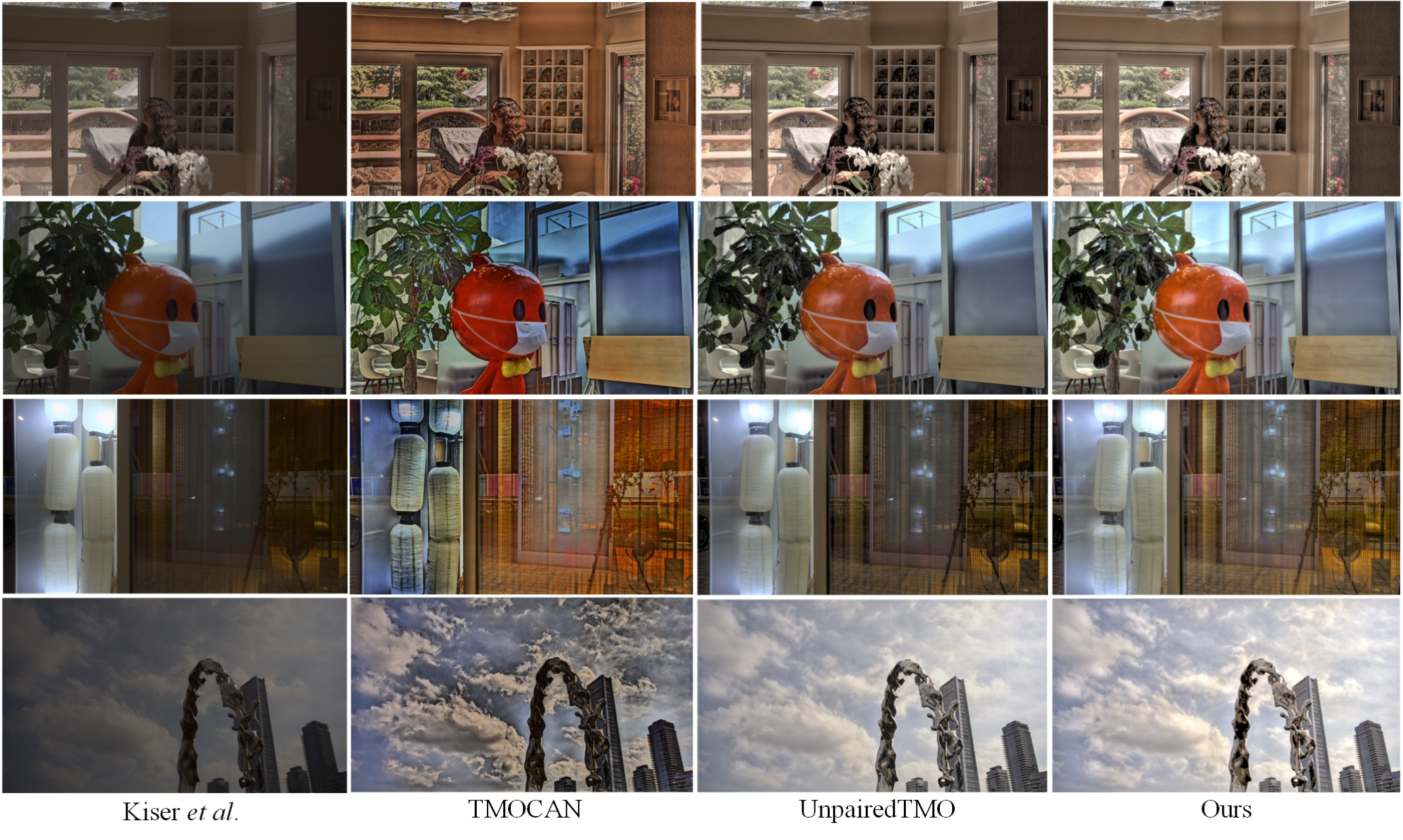}
    \caption{Visual quality comparison for HDR video tone mapping.}
    \label{fig:methodcompare}
\end{figure*}

In this section, we further demonstrate the effectiveness of the proposed TMO for video tone mapping. Since there is no DNN-based video TMOs, we compare with state-of-the-art traditional image \cite{farbman2008edge,liang2018hybrid} and video tone mapping methods \cite{kiser2012real}, DNN-based image tone mapping methods \cite{le2021perceptually,vinker2021unpaired} and image enhancement methods \cite{guo2020zero,jiang2021enlightengan} on our proposed HDR video test set. Among them, the works in \cite{le2021perceptually,guo2020zero} only utilize HDR data for training, namely zero-reference based solution. The works in \cite{vinker2021unpaired,jiang2021enlightengan} and our method utilize unpaired HDR-LDR data for unsupervised learning. For fair comparison, we retrain all the DNN-based methods on our dataset (due to this reason, many tone-mapping methods without codes cannot be compared). Table \ref{Comparison} lists the quantitative results. It can be observed that our proposed method achieves the best TMQI and BTMQI scores.
For TMQI, our method outperforms the second best method UnpairedTMO \cite{vinker2021unpaired} by 0.013 and meanwhile our RWE value is smaller than that of \cite{vinker2021unpaired}. It demonstrates that our method is good at revealing details and keeping temporal consistency.  Compared with image enhancement methods, our method achieves nearly 0.15 and 2 gain for TMQI and BTMQI scores, respectively. This verifies that the enhancement methods are not suitable for tone mapping. Compared with traditional tone mapping methods, our method still achieves obvious gains in terms of TMQI and BTMQI scores. Note that the video tone mapping method \cite{kiser2012real} achieves the best RWE value but its TMQI and BTMQI scores are much worse than other methods. It implies that the lower RWE value of \cite{kiser2012real} may be caused by results with low contrast.


Fig. \ref{fig:methodcompare} presents the visual comparison results on four scenes in our test set. Since the retrained image enhancement methods have uncompetitive results and the page space is limited, we only show the visual comparison results with the video tone mapping method (a) \cite{kiser2012real} and two DNN based tone-mapping methods (b) TMOCAN \cite{le2021perceptually}, and (c) UnpairedTMO \cite{vinker2021unpaired}. Full comparisons are provided in our supplementary file.  It can be observed that our method has the best visual quality with suitable brightness and contrast.  The video tone mapping method \cite{kiser2012real} generates darker results than other methods, but the lantern in the third row is over-exposed. \cite{le2021perceptually} generates over-enhanced results which have unnatural contrast and textures. For example, for the fourth image, the buildings are under-enhanced and the clouds are over-enhanced. UnpairedTMO \cite{vinker2021unpaired} can generate better results, but the wall and curtain in the first and third scenes are under-exposed.  In contrast, our results have pleasant brightness and contrast in all areas.

\subsection{User Study}



\begin{figure}
    \centering
    \includegraphics[width=0.99\linewidth]{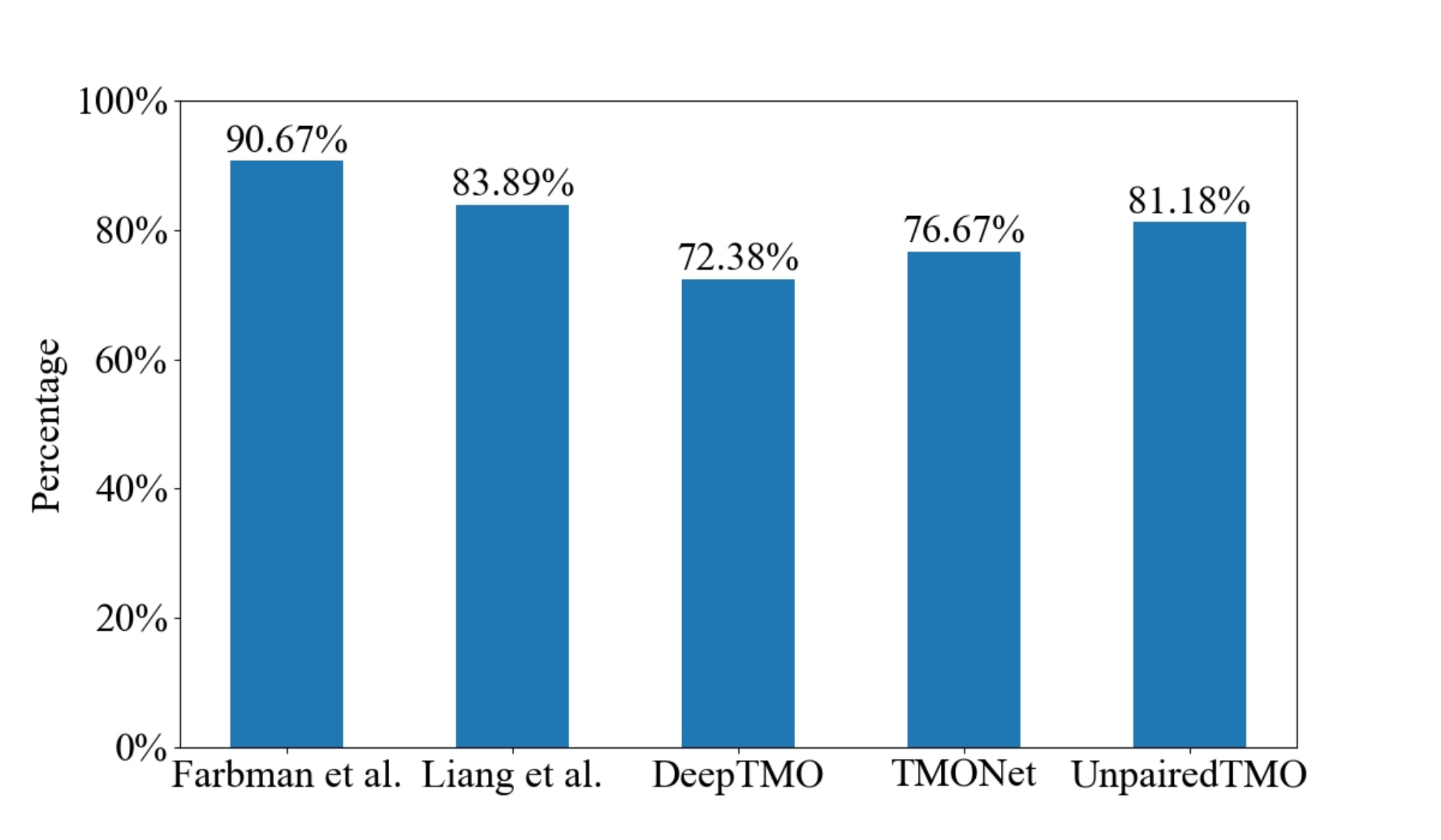}
    \caption{User study for HDR image tone mapping. The Y-axis indicates the percentage of users that prefer our method over other methods.}
    \label{fig:userstudyimg}
\end{figure}

To further verify the superiority of our image TMO results, we conduct a user study by comparing our method with five competitive methods, including \cite{farbman2008edge,liang2018hybrid}, DeepTMO \cite{rana2019deep}, TMONet \cite{panetta2021tmo}, and UnpairedTMO \cite{vinker2021unpaired}. For each comparison, the users are asked to select one image with better visual quality from two images, among which one is generated by our method. The visual quality includes contrast, naturalness, and detail preserving. There are 15 subjects involved in the user study and we randomly select 30 scenes from our test set for evaluation. As shown in Fig. \ref{fig:userstudyimg}, our method outperforms all the compared five methods.

\begin{figure}
    \centering
    \includegraphics[width=0.8\linewidth]{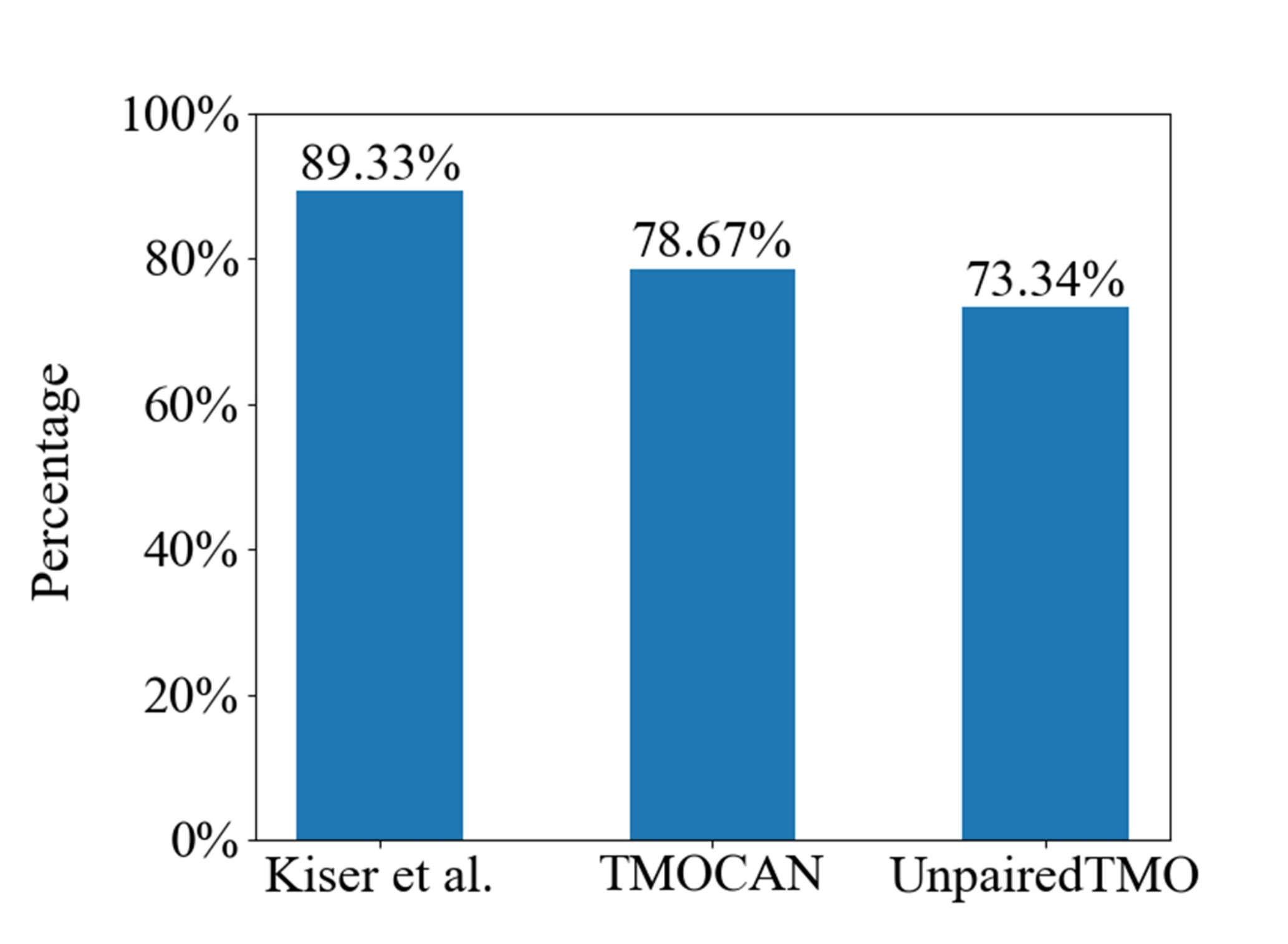}
    \caption{User study for HDR video tone mapping. The Y-axis indicates the percentage of users that prefer our method over other methods.}
    \label{fig:userstudyvideo}
\end{figure}

We also perform user study for video TMO quality evaluation by comparing our method with three competitive methods, including \cite{kiser2012real}, TMOCAN \cite{le2021perceptually}, and UnpairedTMO \cite{vinker2021unpaired}. Similar to the user study for image TMO evaluation, we ask users to select one video with better quality from two videos which includes our result. Besides the image quality measurement, the video quality is further measured by temporal consistency. There are totally 15 subjects and 10 videos are included in the user study. As shown in Fig. \ref{fig:userstudyvideo}, users prefer our method over the compared methods.

\subsection{Computing Complexity}
In this section, we further give the computing complexity comparison in terms of multiply-add operations (MACs).
As shown in Table VI, we list the MACs required for each method to generate one LDR frame output with a resolution of 256$\times$256.
For our method, we give three realizations with different channel numbers. Specifically, we reduce the channel number of IVTMNet$_{\text{full}}$ to construct IVTMNet$_{0.75}$ and IVTMNet$_{0.5}$, whose channel numbers are 0.75 and half of our full solution, respectively. It can be observed that IVTMNet$_{\text{full}}$, IVTMNet$_{0.75}$ and IVTMNet$_{0.5}$ are the top three models in terms of TMQI and BTMQI scores. IVTMNet$_{0.5}$ has the lowest MACs and its RWE value is slightly worse than that of IVTMNet$_{\text{full}}$. IVTMNet$_{0.75}$ achieves a good balance between tone mapping performance and computing complexity.

\section{Conclusion}

In this paper, we explore unsupervised image and video tone mapping by proposing an effective IVTMNet and a set of unsupervised loss functions. It is convenient to switch between image and video TMO by introducing TFR module, which can improve the temporal consistency of video tone-mapped results. To improve unsupervised training, we propose domain and instance based contrastive learning loss and propose a new latent space in which to measure the similarities in negative and positive pairs. Experimental results on image and video tone mapping dataset demonstrate the superiority of the proposed method in enhancing brightness and contrast and keeping temporal consistency of tone-mapped results. 


Our work also has some limitations. First, our network consumes a significant amount of computing resources compared with traditional TMOs. In the future, we would like to reduce the computing cost by designing more computation-friendly layers. Second, HDR images cover many lighting moods, from low key with high contrast to high key. However, during TMO, we did not design loss functions to help preserve the lighting moods of tone mapped results. We expect more works to be conducted on this topic.

%

%

\ifCLASSOPTIONcaptionsoff
  \newpage
\fi



\bibliographystyle{IEEEtran}
\bibliography{egbib}
%
%

%




\end{document}